\documentclass[letterpaper]{article}
\usepackage[preprint]{aaai2027}
\usepackage[hyphens]{url}  
\usepackage{graphicx} 
\usepackage{natbib}  
\usepackage{caption} 
\usepackage{algorithm}
\usepackage{algorithmic}

\usepackage{booktabs}
\usepackage{array}

\title{AsmEvo: Agentic Assembly-Level Optimization of AMD GPU Kernels with Functional Equivalence Verification}

\author{
\normalsize
Ji Liu\textsuperscript{\rm 1},
Puyuan Yang\textsuperscript{\rm 1},
Rongzhang Zheng\textsuperscript{\rm 1},
Fan Wang\textsuperscript{\rm 1},
Jinglin Wang\textsuperscript{\rm 1,\rm 2}\thanks{Work performed during an internship at AMD.},
Muhammad A. Awad\textsuperscript{\rm 1},
Mortis Huang\textsuperscript{\rm 1},
Andy Chang\textsuperscript{\rm 1},
Zekai Li\textsuperscript{\rm 1},
Zeping Li\textsuperscript{\rm 1},
Zihao An\textsuperscript{\rm 1},\\
Yue Liu\textsuperscript{\rm 1},
Yuchen Yang\textsuperscript{\rm 1},
Jianghui Wang\textsuperscript{\rm 1},
Chushi Chen\textsuperscript{\rm 1},
Ziqiong Liu\textsuperscript{\rm 1},
Fuwei Yang\textsuperscript{\rm 1},
Dong Li\textsuperscript{\rm 1},
Wen Heng Chung\textsuperscript{\rm 1},
Shengcai Liu\textsuperscript{\rm 2},
Emad Barsoum\textsuperscript{\rm 1}
}

\affiliations{
\textsuperscript{\rm 1}Advanced Micro Devices, Inc. (AMD)\\
\textsuperscript{\rm 2}Southern University of Science and Technology, Shenzhen, China
}

\begin{document}

\maketitle

\begin{abstract}
High-performance ML systems increasingly rely on GPU kernels whose editable source is unavailable, generated, or too distant from final machine code to expose remaining optimizations. Existing LLM kernel optimizers and autotuners mainly operate on CUDA, Triton, HIP, or tensor-program source and validate against reference implementations. We study a stricter setting: optimizing an already compiled AMDGPU code object, where the deployed binary is the only behavioral oracle.

We present AsmEvo, an agentic assembly-level optimizer for AMD GPU kernels. Given an AMDGPU code object $K_0$, AsmEvo reconstructs a reassemblable representation, proposes low-level edits with a long-horizon agent, rebuilds an ABI-preserving optimized object, and accepts candidates only after differential verification against $K_0$ under identical launches. AsmEvo combines code-object recovery, metadata-aware rebuilding, profiling-guided hot-window editing, correctness-gated timing, and conservative in-place patch fallback.

We conduct extensive experiments with AsmEvo on various AMD GPU kernels. On MI308X, AsmEvo improves 29 of 30 selected KernelBench kernels, reaching 1.35x geometric-mean and 3.88x maximum speedup. On MI300X production workloads, it improves all evaluated AITer binaries and vLLM/SGLang Triton assembly kernels, reaching 1.09x/1.31x and 1.18x/1.34x geometric-mean/maximum speedups, respectively, while preserving functional equivalence.
\end{abstract}

\section{Introduction}
Modern machine learning workloads depend on GPU kernels whose final performance is decided
not only by high-level schedules, but also by low-level decisions made after lowering:
wait-counter placement, instruction selection, register allocation, kernel descriptors, and
ABI-visible metadata~\cite{tillet2019triton,tvm2018,ansor2020}. Compilers~\cite{tvm2018}, tensor-program autotuners~\cite{ansor2020,chen2018learning}, DSLs such as Triton, and vendor
libraries such as AITer~\cite{aiter}, rocBLAS~\cite{rocblas}, and Composable Kernel~\cite{composablekernel} are highly effective, yet they also
create a practical boundary. Once a kernel has become an AMDGPU code object~\cite{amdgpu_code_object}, the remaining
optimization opportunities may no longer be exposed through the source program---and in many
deployments the source is not available at all.

This source-free setting is common in deployed ML systems: serving stacks may expose only a
compiled HSACO~\cite{amdgpu_code_object}, a JIT cache artifact~\cite{tillet2019triton}, or a vendor binary, such as AITer code objects or
Triton-generated kernels embedded in vLLM~\cite{kwon2023efficient} and SGLang~\cite{zheng2024sglang}. Re-optimizing the deployed artifact is
attractive because it targets the exact code that runs on the GPU, but it is brittle: a local
instruction edit can corrupt memory, violate the kernarg ABI, desynchronize metadata, or
appear faster only by changing semantics~\cite{nvbit2019,raayai2025luthier}.

Recent LLM-based kernel systems~\cite{wang2025geak,zhang2025cudaforge,dai2026cuda,chen2026avo} generate and refine CUDA, Triton, HIP, or tensor-program
kernels from high-level specifications, including KernelBench \citep{kernelbench2025}. They
assume an editable source artifact and an independent reference implementation. Compiled
code-object kernels provide neither: the original binary is both the deployed artifact and
the only behavioral oracle.

This paper asks whether a kernel can be optimized \emph{directly at the assembly level}
using only the original compiled binary to define acceptable behavior. We present AsmEvo, a
post-compilation optimizer for AMDGPU code objects and recovered AMDGCN assembly. AsmEvo
reconstructs a reassemblable representation, exposes hot instruction windows to a search
driver, rebuilds edited candidates while preserving the launcher ABI, and accepts a candidate
only after differential verification against the original binary under identical inputs and
launch configurations. Thus the optimization signal is non-hackable: fast but incorrect
candidates are rejected before timing.

The central challenge is obtaining a faithful oracle from a binary with no high-level
signature. AsmEvo uses a two-tier differential oracle. For kernels whose launch structure is
inferable from metadata, it synthesizes guarded inputs and compares outputs directly. For
mixed-dtype, strided, block-table, or pointer-rich production kernels, it captures a real
dispatch---kernargs, launch geometry, and referenced device memory---and replays candidates
against that state using tolerance-aware checks for floating-point outputs and byte-exact
checks for integer and opaque state.

AsmEvo turns verified edits into loadable objects through a metadata-aware rebuild route: it regenerates descriptors and AMDGPU metadata while freezing the kernarg layout, so the optimized object remains a drop-in replacement. A conservative in-place byte patch is used only when rebuild fails and the edit is resource-neutral and size-non-increasing. Profiling and static analysis localize hot instruction windows, which are edited, spliced back into the full assembly, rebuilt, and reverified.

On top of this environment, AsmEvo uses a long-horizon agentic search driver under external verification. Deterministic gates own build validation, resource checks, equivalence, timing, commit thresholds, and lineage management; the agent plans architecture-specific edits, uses failed attempts as memory, explores diverse verified start points, composes compatible improvements, and redirects stalled searches. This separation lets learning-based exploration help where useful while every accepted result is certified externally.

We conduct extensive experiments with AsmEvo on diverse AMD GPU kernels. On MI308X, AsmEvo improves 29 of 30 selected KernelBench kernels, reaching 1.35x geometric-mean and 3.88x maximum speedup. On MI300X production workloads, it improves all evaluated AITer binaries and vLLM/SGLang Triton assembly kernels, reaching 1.09x/1.31x and 1.18x/1.34x geometric-mean/maximum speedups, respectively, while preserving functional equivalence. These results show that AsmEvo complements source-level compilers, JIT systems, autotuners, and production inference libraries.

Our contributions are as follows:
\begin{itemize}
  \item We formulate \textbf{source-free, ABI-preserving GPU kernel optimization} as a correctness-gated post-compilation problem over AMDGPU code objects, where the original binary is the differential correctness oracle and every reported speedup must pass external verification.

  \item We build a \textbf{binary recovery, rebuild, and differential-verification pipeline}: round-trip recovery of reassemblable AMDGCN assembly, metadata-aware ABI-preserving rebuild, conservative patch fallback, and a two-tier oracle combining synthetic launch inference with real-dispatch replay.

  \item We instantiate a \textbf{long-horizon agentic search driver under external verification}: the agent plans architecture-specific edits, uses failed attempts as memory, explores diverse verified start points, composes compatible improvements, and recovers from stalls, while deterministic gates own correctness and timing. Evaluations on KernelBench, AITer, and vLLM/SGLang Triton JIT kernels show consistent post-compilation gains.
\end{itemize}

\section{Related Work}

\paragraph{LLM Kernel Optimization and Tensor Compilers.}
LLM-based kernel optimization has developed along several complementary directions.
KernelBench~\citep{kernelbench2025} established an execution-guided setting in which generated kernels are iteratively
refined using correctness and performance feedback. Subsequent work
explores reinforcement learning with measured rewards
\citep{kevin2026,cudal12026,drkernel2026,dai2026cuda}, agentic and evolutionary search over candidate
implementations \citep{kernelevolve2026,robustkbench2025}, and specialized kernel-generation
models \citep{kernelllm2025}. Related benchmarks~\citep{tritonbench2025,robustkbench2025} further show that performance optimization
remains difficult and that insufficient validation can reward fast but semantically incorrect
kernels. In parallel, tensor compilers and DSLs
optimize schedules, tiling, memory movement, and code generation, including Halide
\citep{halide2013}, TVM \citep{tvm2018}, AutoTVM \citep{chen2018learning}, Ansor
\citep{ansor2020}, MetaSchedule \citep{metaschedule2022}, and Triton
\citep{tillet2019triton}. Despite their different search mechanisms, these approaches operate
before or during compilation and assume editable source or IR together with a high-level
correctness reference. AsmEvo targets the complementary post-compilation setting: its input
is the emitted AMDGPU code object, and the original binary serves as the behavioral oracle.


\paragraph{Assembly Optimization and Binary Rewriting.}
Superoptimizers search for faster programs at the instruction or IR level. STOKE
\citep{stoke2013} applies stochastic search to x86 assembly, Souper
\citep{souper2017} discovers missing LLVM optimizations, and AlphaDev
\citep{alphadev2023} learns low-level sorting routines. Recent learning-based systems further
optimize assembly code: \citet{asmrl2025} apply reinforcement learning to CPU assembly,
while CuAsmRL~\citep{he2025cuasmrl}   formulates NVIDIA SASS instruction scheduling as an
assembly game and learns to reorder instructions for higher throughput. These methods assume
CPU-oriented specifications or operate on NVIDIA SASS schedules with constrained
transformations. AsmEvo instead targets source-free AMDGPU code objects and jointly addresses
assembly recovery, ABI- and metadata-preserving rebuild, broader agent-proposed edits, and
differential verification against the original binary.

GPU binary frameworks primarily target instrumentation. NVBit \citep{nvbit2019} modifies
NVIDIA SASS, while Luthier \citep{raayai2025luthier} operates on loaded ROCm code objects for
instrumentation and analysis. Added overhead is acceptable for these uses, whereas AsmEvo
seeks faster, verified drop-in objects while preserving launcher semantics, resource
declarations, descriptors, and metadata.

\paragraph{Correctness of Optimized Code.}
Compiler and translation-validation systems check whether optimizations preserve semantics;
Alive2~\citep{alive22021}, for example, validates LLVM transformations. Program-equivalence
reasoning remains difficult even for strong language models~\citep{equibench2025}, motivating
execution-based verification rather than model judgment. Existing GPU optimizers typically
compare against source-level or framework references. Such references are unavailable for
HSACO-only artifacts, so AsmEvo treats the original binary as a differential oracle and
checks equivalence before timing. Its guarantees are empirical and limited to the evaluated
inputs and launch configurations.

\paragraph{Positioning of AsmEvo.}
To our knowledge, AsmEvo is the first system to combine agentic search, ABI- and
metadata-preserving AMDGPU code-object rebuilding, and differential verification against the
original deployed binary. Existing source-level kernel agents are not designed to consume
HSACO-only artifacts or certify drop-in equivalence without this recovery and verification
harness. We therefore compare search drivers within the same harness, holding recovery,
rebuilding, correctness verification, and commit gates fixed. AsmEvo complements rather than
replaces source-level compilers, autotuners, and inference libraries.

\section{Method}
\label{sec:method}

\subsection{Overview}
\label{sec:overview}
AsmEvo optimizes compiled AMDGPU code objects without source code or an independent reference
implementation. It treats the original object $K_0$ as a differential oracle and admits an
optimized object $K'$ only after ABI-preserving rebuild, functional-equivalence verification,
and performance measurement. We denote the recovered assembly by $s(K)$, evaluated inputs by
$\mathcal{X}$, and verified speedup by $\hat{s}(K)=T(K_0)/T(K)$. A deterministic controller
owns all correctness-critical operations, while a modular long-horizon LLM driver proposes
edits and replans from structured feedback. This separation creates a verifiable environment
in which fast but incorrect candidates cannot advance the search.

\begin{figure*}[t]
\centering
\includegraphics[width=1.0\textwidth]{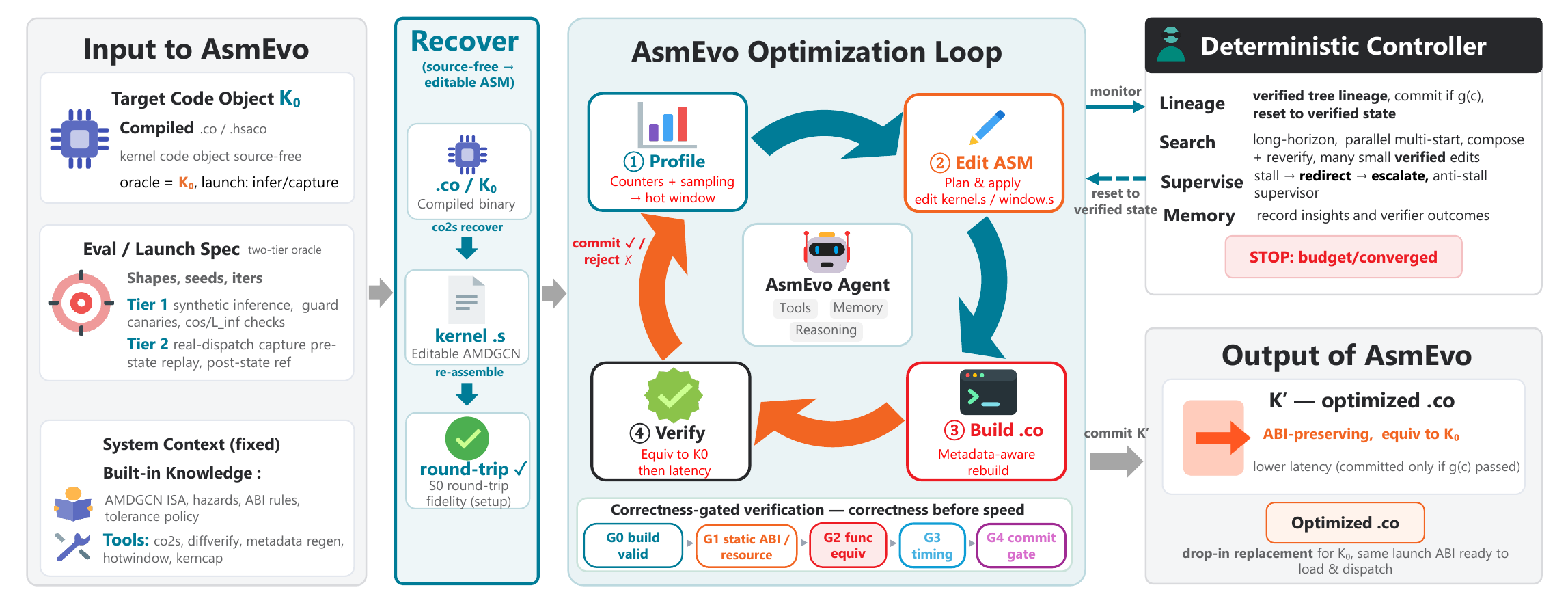}
\caption{AsmEvo combines faithful code-object recovery, metadata-aware ABI-preserving rebuild,
and correctness-before-performance evaluation against the original $K_0$. A modular LLM
driver proposes edits through multi-start search, verified composition, and anti-stall
redirection, while only verified, latency-improving candidates enter the lineage.}
\label{fig:arch}
\end{figure*}

\subsection{Problem Formulation}
\label{sec:formulation}
Given $K_0$, AsmEvo seeks an optimized code object $K'$ satisfying three requirements.

\begin{description}
\item[R1 (ABI preservation).]
$K'$ preserves symbols, kernarg layout, launch semantics, and externally visible metadata,
allowing the original host launcher to invoke it unchanged.

\item[R2 (Functional equivalence).]
For every evaluated input $x\in\mathcal{X}$, $K'$ matches the output of the original-binary
oracle $O(x)=\mathrm{run}(K_0,x)$:
\begin{equation}
\begin{array}{l}
\mathrm{eq}(K',x)\equiv
c_x\ge\theta \,\wedge\,
\|\Delta_x\|_\infty\le\tau \,\wedge\, \iota(K',x),\\
c_x=\cos\!\big(K'(x),O(x)\big),\qquad
\Delta_x=K'(x)-O(x),
\end{array}
\label{eq:equiv}
\end{equation}
where $\iota$ requires bit-exact integer buffers and intact out-of-bounds guards.

\item[R3 (Latency improvement).]
$K'$ reduces measured latency. Commits are judged against the current verified best, while
reported speedup is $\hat{s}(K')=T(K_0)/T(K')$ relative to the original object.
\end{description}

Equivalence is empirical rather than formal and is scoped to the evaluated inputs and launch
configurations. Unlike source-level autotuning, AsmEvo optimizes an already compiled binary,
preserves its ABI, and verifies behavior against the unmodified deployed object.

\subsection{Code-Object Recovery and Round-Trip Fidelity}
\label{sec:recovery}
AsmEvo reconstructs editable assembly $s(K_0)$ from the AMDGPU ELF container. Because raw
disassembly omits the declaration layer, AsmEvo recovers sections, symbols, notes, kernel
descriptors, AMDGPU metadata, and the AMDGCN instruction body. It also reconstructs
\texttt{.amdhsa\_kernel} declarations and re-symbolizes PC-relative control flow so that
branches remain valid after instruction insertion, deletion, or reordering.

A one-time round-trip gate reassembles and relinks $s(K_0)$ and compares the regenerated
instruction body, kernel descriptors, and metadata with $K_0$, masking only fields determined
by linking. Passing this gate establishes that later edits operate on a faithful, rebuildable
representation. Rare recovery repairs are accepted only when the repaired representation
passes the same byte-level round-trip check.

\subsection{Input Acquisition and Differential Oracle}
\label{sec:oracle}
Equivalence requires executing $K_0$ and each candidate with identical inputs and launch
specifications. However, a code object exposes neither a high-level function signature nor
the application logic that constructs pointer-rich arguments. AsmEvo therefore provides a
two-tier oracle acquisition mechanism.

\paragraph{Tier 1: synthetic inference.}
When launch structure is recoverable from descriptors and metadata, AsmEvo infers argument
offsets, scalar types, pointer roles, buffer sizes, and launch dimensions. It then generates
deterministic inputs covering representative shapes, strides, dtypes, boundary values, and
random seeds. Allocations include guard regions initialized with canaries, allowing the
equivalence check to detect out-of-bounds writes in addition to output divergence.

\paragraph{Tier 2: real-dispatch capture.}
Metadata alone is insufficient for many production kernels, including those using mixed
dtypes, non-contiguous tensors, block tables, nested pointer structures, or application-defined
workspace layouts. For such kernels, AsmEvo intercepts a real application dispatch and records
its kernarg buffer, grid and workgroup dimensions, referenced device-memory regions, and
pre-dispatch memory state. The original object is executed once to produce the reference
post-state.

Candidates replay the captured dispatch with identical launch parameters and memory contents.
When captured arguments contain absolute device pointers, memory regions are restored at their
original virtual addresses so that nested pointers remain valid without reconstructing
application-level data structures. Floating-point outputs are checked using
Eq.~\ref{eq:equiv}, while integer buffers, opaque state, and guard regions are checked
byte-exactly. This mechanism allows AsmEvo to verify deployed kernels even when no standalone
source-level test harness is available.

\subsection{Metadata-Aware Rebuild}
\label{sec:rebuild}
AsmEvo converts each edited assembly into a loadable code object by rescanning it for
VGPR/SGPR/AGPR, LDS, and scratch usage. It then recomputes resource fields in both the kernel
descriptor and AMDGPU metadata before relinking the object. The kernarg layout, kernarg
segment size, symbol interface, and launch semantics remain frozen, preserving compatibility
with the original launcher.

If normal rebuilding fails because an object contains unsupported or incompletely recovered
constructs, AsmEvo may use a conservative in-place byte patch. This fallback is restricted to
resource-neutral and size-non-increasing edits, retaining the original descriptors and
metadata. Patched candidates still pass static consistency, differential equivalence, and
performance gates; the fallback therefore changes only the rebuild mechanism, not the
acceptance criterion.

\subsection{Profiling and Hot-Window Localization}
\label{sec:profile}
Hardware counters, instruction sampling, and static analysis identify stall-dominant
instruction windows in large kernels. The resulting profile summarizes instruction mix,
dependency chains, memory behavior, occupancy constraints, and resource pressure for the
search driver. Instead of placing the entire assembly in every optimization context, AsmEvo
selects a hot window together with the surrounding data dependencies and relevant metadata.

A proposed local edit is spliced back into the complete assembly before whole-kernel rebuild
and verification. Hot-window localization therefore reduces search context and discourages
unrelated modifications without weakening the full-kernel acceptance criterion.

\subsection{Gated Verification Harness}
\label{sec:gates}
AsmEvo performs two setup checks once per kernel: round-trip fidelity and oracle acquisition.
Each candidate then passes Algorithm~\ref{alg:verify}, which checks assembly validity, static
resource and ABI consistency, functional equivalence, and finally performance. Correctness
strictly precedes timing, preventing broken but fast candidates from receiving a positive
optimization signal.

Timing uses warmup followed by the median of $R$ event-timed launches. Candidate timing and
oracle execution occur under the same device and launch conditions. Distinct outcomes such
as assembly failure, metadata inconsistency, output divergence, runtime failure, and
insufficient speedup are returned as structured feedback to the search driver.

\begin{algorithm}[t]
\caption{\textsc{GatedEval}: verify and time a candidate edit.}
\label{alg:verify}
\begin{algorithmic}[1]
\REQUIRE edited assembly $s'$; oracle $(\mathcal{X},O)$; original $K_0$
\IF{$\neg\textsc{Assembles}(s')$}
  \RETURN $\langle\textsc{asm\_invalid}\rangle$
\ENDIF
\STATE $c \gets \textsc{Build}(s')$
\IF{$\neg\textsc{StaticConsistent}(c)$}
  \RETURN $\langle\textsc{abi\_invalid}\rangle$
\ENDIF
\FOR{$x \in \mathcal{X}$}
  \IF{$\neg\,\mathrm{eq}(c,x)$}
    \RETURN $\langle\textsc{divergent}\rangle$
  \ENDIF
\ENDFOR
\STATE $\hat{s}(c)\gets T(K_0)/T(c)$
\RETURN $\langle\textsc{ok},\hat{s}(c)\rangle$
\end{algorithmic}
\end{algorithm}

\subsection{Long-Horizon Agentic Search}
\label{sec:search}
The verification harness accepts any candidate generator; AsmEvo instantiates it with a
long-horizon LLM search driver. A deterministic controller owns the wall-clock budget,
workspace state, rebuild and verification calls, commit decisions, candidate lineage, and
termination conditions. The LLM analyzes profiling evidence, selects bottlenecks, proposes
localized assembly transformations, interprets structured failures, and replans after
unproductive attempts. Thus, the model may guide exploration, but it cannot declare either
correctness or performance.

\paragraph{Commit gate and verified lineage.}
A candidate $c$ is committed only if
\begin{equation}
\begin{array}{c}
m^\star=\big(1+\max(\epsilon,k\!\cdot\!\mathrm{cv})\big)\hat{s}(K^\star),\\[2pt]
g(c)\equiv
\Big[\textstyle\bigwedge_x\mathrm{eq}(c,x)\Big]\wedge
\hat{s}(c)\ge\max(m^\star,s_{\mathrm{floor}}),
\end{array}
\label{eq:gate}
\end{equation}
where $\mathrm{cv}$ is the timing coefficient of variation, $\epsilon$ is a minimum
improvement margin, and $s_{\mathrm{floor}}$ rejects trivial rewrites. The variance-aware
threshold prevents measurement noise from advancing the search.

Accepted candidates form a tree-structured lineage rather than a single destructive chain.
Each node records its parent, verified speedup, changed instruction windows, resource usage,
and optimization rationale. Workers may therefore branch from the current best or return to
an earlier verified version when a promising direction reaches a local optimum. The reported
global best follows a no-regress invariant.

\paragraph{Multi-start exploration.}
In single mode, the driver proposes one edit from the current best. In team mode, a planner
assigns orthogonal optimization directions---such as latency hiding, dependency reduction,
register-pressure control, memory-access restructuring, or instruction simplification---to
parallel workers. Start points are selected from diverse verified lineage nodes rather than
requiring every worker to modify the same candidate. Every proposal is independently rebuilt
and passed through \textsc{GatedEval}.

\paragraph{Verified composition.}
Individually successful edits are not assumed to remain correct or beneficial when combined.
The integrator considers candidates that share a base and modify disjoint or complementary
instruction regions. It applies their edits incrementally, rebuilding and verifying after
each step. A composition is committed only if it passes all correctness gates and improves
over both parent candidates; otherwise, the best individual candidate is retained.

\paragraph{Memory and anti-stall supervision.}
A bounded run memory stores successful optimization insights, structured gate outcomes,
failure signatures, and hashes of repeated candidates. If the controller detects prolonged
no-improvement, repeated proposals, or infrastructure failures, it requests a redirect that
summarizes attempted directions and marks exhausted hypotheses. The driver may then change
the targeted bottleneck or backtrack to an earlier verified node. Persistent stalls escalate
to wider multi-start exploration. All redirected candidates remain subject to the same
deterministic gates.

Algorithm~\ref{alg:search} summarizes the complete loop. The controller resets each worker
to a verified state after evaluation, preventing unverified modifications from accumulating
across iterations.

\begin{algorithm}[t]
\caption{AsmEvo long-horizon verified search.}
\label{alg:search}
\begin{algorithmic}[1]
\REQUIRE original object $K_0$; budget $B$; directions $M$; GPUs $G$
\STATE \textbf{assert} \textsc{RoundTrip}$(s(K_0),K_0)$
\STATE $(\mathcal{X},O)\gets\textsc{AcquireOracle}(K_0)$
\STATE $K^\star\gets K_0$; initialize lineage $\mathcal{L}$ and memory $\mathcal{M}$
\WHILE{elapsed time $<B$}
  \STATE $D\gets\textsc{Plan}(K^\star,\mathrm{profile},\mathcal{M},M)$
  \STATE $P\gets\textsc{StartPoints}(\mathcal{L},D)$; $C\gets\emptyset$
  \FOR{each $(d,b)\in P$ assigned to GPU $g\in G$}
    \STATE $s'\gets\textsc{Edit}(s(b),d,\mathcal{M})$
    \STATE $r\gets\textsc{GatedEval}(s',(\mathcal{X},O),K_0)$
    \IF{$r=\langle\textsc{ok},\hat{s}\rangle$}
      \STATE $C\gets C\cup\{(s',b,\hat{s})\}$
    \ENDIF
  \ENDFOR
  \STATE $\tilde{c}\gets\textsc{Integrate}(C)$
  \IF{$\tilde{c}\neq\emptyset$}
    \STATE $r^\star\gets\textsc{GatedEval}(s(\tilde{c}),(\mathcal{X},O),K_0)$
    \IF{$r^\star=\langle\textsc{ok},\hat{s}^\star\rangle \wedge g(\tilde{c})$}
      \STATE $K^\star\gets\tilde{c}$; append to $\mathcal{L}$; re-profile
    \ENDIF
  \ENDIF
  \STATE $\mathcal{M}\gets\textsc{UpdateMemory}(\mathcal{M},C)$
  \IF{stalled}
    \STATE \textsc{RedirectOrEscalate}$(\mathcal{L},\mathcal{M})$
  \ENDIF
  \IF{converged}
    \STATE \textbf{break}
  \ENDIF
  \STATE reset worker workspaces to verified $K^\star$
\ENDWHILE
\RETURN $K^\star$ and lineage $\mathcal{L}$
\end{algorithmic}
\end{algorithm}

\subsection{Scope and Limitations}
\label{sec:scope}
AsmEvo provides empirical differential equivalence rather than formal verification; its
guarantee is limited to evaluated inputs and launch configurations. Additional shapes require
new inferred inputs or dispatch captures. The overall harness is architecture-agnostic, but
the current recovery and metadata-aware rebuild implementation is calibrated for CDNA-class
AMD GPUs. Complex application state may require real-dispatch capture, and achievable gains
depend on optimization headroom remaining in the compiled kernel.

\section{Experiments}

We evaluate whether AsmEvo (1) improves compiled benchmark kernels and (2) transfers to
source-free production artifacts under a fixed verification harness. We use KernelBench Level~1 and Level~2, AITer code objects, and
Triton JIT HSACOs emitted by vLLM and SGLang. The unmodified binary is always the
$1.00\times$ reference.

\begin{table}[t]
\centering\small
\setlength{\tabcolsep}{3pt}
\begin{tabular}{@{}lcp{0.56\columnwidth}@{}}
\toprule
Source & Size & Coverage \\
\midrule
KernelBench L1 & 15 & activations, losses, reductions/norms, depthwise conv, matmul, pooling \\
KernelBench L2 & 15 & fused GEMM/matmul and conv pipelines, normalization, activations, pooling, reductions \\
AITer & 4 & code-object MoE, FP8 FMHA, BF16 FMHA, BF16 FMHA backward \\
Triton JIT & 4 & vLLM/SGLang A8W8 GEMM, MoE, MoE LoRA, AWQ GEMM \\
\bottomrule
\end{tabular}
\caption{Kernel sources, counts, and operator coverage.}
\label{tab:bench}
\end{table}

\subsection{Experimental Setup}
Experiments use CDNA3 AMD Instinct GPUs with ROCm: KernelBench Level~1 and Level~2 run on
MI308X, while AITer and the Triton JIT HSACOs run on MI300X. All search agents use
Claude Opus 4.8 with provider-default decoding. By default, AsmEvo performs multi-start
search with four GPU workers under a fixed 0.5-day (12-hour) wall-clock budget per kernel.
For each kernel, AsmEvo recovers a reassemblable AMDGCN representation, performs a one-time
round-trip fidelity check, and builds the differential oracle. Every candidate passes
\textsc{GatedEval} in a fixed order: functional-equivalence verification first, performance
measurement second. The equivalence check is repeated three times per candidate, using
$\theta=0.9999$ and $\tau=10^{-3}$ in Eq.~\ref{eq:equiv}. Latency is the median of event-timed
launches after warmup. Because the hardware is shared and clocks cannot be locked, we use
a large, representative launch configuration to suppress launch and event noise. Commits
use the variance-aware margin in Eq.~\ref{eq:gate}, with $\epsilon=0.002$, $k=0.85$,
and $s_{\mathrm{floor}}=1.0$. We report speedups rather than raw latencies in all result
tables.

\begin{table*}[t]
  \centering\small
  \setlength{\tabcolsep}{4pt}
  \begin{tabular}{@{}lrrrrrrll@{}}
  \toprule
  Split & Kernels & Improved & Geo. & Max & Attempts/K & Commits/K & GPU & Best case \\
  \midrule
  KernelBench L1 & 15 & 15 (100\%) & 1.56 & 3.88 & 119.3 & 7.2 & MI308X & Depthwise Conv2D \\
  KernelBench L2 & 15 & 14 (93\%) & 1.17 & 2.55 & 156.4 & 6.8 & MI308X & Matmul+Swish+Sum+GroupNorm \\
  All KernelBench & 30 & 29 (97\%) & 1.35 & 3.88 & 137.8 & 7.0 & MI308X & Depthwise Conv2D \\
  \bottomrule
  \end{tabular}
  \caption{KernelBench performance under the default AsmEvo configuration. Speedups are measured against the original compiled code object and reported only after functional-equivalence verification.}
  \label{tab:kbench-results}
\end{table*}

\subsection{Artifact Sources and Coverage}
Table~\ref{tab:bench} lists the four sources. They cover the operator types most common in large language models: 15 validated KernelBench Level~1 operators, 15 KernelBench Level~2 operators, and 8 production kernels comprising 4 AITer code objects and 4 Triton JIT HSACOs.

The sources differ in how the input binary is obtained. For KernelBench, we use an AMD GEAK-style agent~\cite{wang2025geak} with GPT-5.0 to refine HIP versions of the PyTorch benchmark kernels for five rounds, compile them, and give AsmEvo only the compiled object as $K_0$. Production kernels are used exactly as deployed. We select AITer code objects from MI300X workloads and collect the Triton JIT HSACOs emitted by vLLM and SGLang during inference. Each binary is paired with a captured real dispatch---kernargs, launch geometry, and referenced device memory---for deterministic replay. The capture is required because these kernels use mixed dtypes and pointer-rich layouts that synthetic launch inference cannot reconstruct, motivating the real-dispatch oracle. The selected Level~1 and Level~2 HIP inputs are all faster than their
PyTorch operators in the source-level benchmark, so $K_0$ is not a weak
baseline.

\begin{table}[t]
  \centering\small
  \setlength{\tabcolsep}{3pt}
  \begin{tabular}{c>{\centering\arraybackslash}p{0.62\columnwidth}c}
  \toprule
  Source & Kernel & Speedup \\
  \midrule
  AITer & \shortstack{aiter\_fmoe\_fp8\_blockscale\\subgu256} & 1.310 \\
  AITer & aiter\_fmha\_hd128\_fp8\_causal & 1.007 \\
  AITer & \shortstack{new\_aiter\_fmha\_hd192x128\\bf16\_causal} & 1.025 \\
  AITer & \shortstack{new\_aiter\_fmha\_bwd\_hd128\\bf16\_causal\_a32} & 1.049 \\
  vLLM & gemm\_a8w8\_blockscale & 1.234 \\
  SGLang & \shortstack{sglang\_fused\_moe\\lora\_kernel} & 1.025 \\
  vLLM & fused\_moe\_kernel & 1.343 \\
  SGLang & awq\_gemm\_kernel & 1.133 \\
  \bottomrule
  \end{tabular}
  \caption{Verified per-kernel speedups over the original production binaries
on MI300X. Only binary-level AITer and assembly-level Triton results are included.
The family geometric means are $1.09\times$ for AITer and
$1.18\times$ for Triton JIT. Triton JIT HSACOs are collected from vLLM
and SGLang and evaluated using captured production dispatches.}
  \label{tab:production-results}
\end{table}

\subsection{Metrics}
For each optimized kernel, verified speedup is
$\hat{s}(K')=T(K_0)/T(K')$, where $K_0$ is the original code object and
$T(\cdot)$ is the median post-warmup latency. For KernelBench splits, we report
geometric-mean and maximum speedups, the number of improved kernels, and the mean
attempts and commits per kernel. For each production-kernel family, we report
the geometric-mean speedup together with individual per-kernel speedups.
A kernel is improved only if its final candidate passes Eq.~\ref{eq:gate} and
has $\hat{s}(K')>1$. The commit rate is the total number of verified commits
divided by the total number of candidates submitted to \textsc{GatedEval}.
Every reported speedup passes functional-equivalence verification before timing.

\subsection{Results}
\subsubsection{KernelBench Kernels}
Table~\ref{tab:kbench-results} reports the default AsmEvo results on MI308X.
AsmEvo improves all 15 KernelBench Level~1 kernels, reaching a
$1.56\times$ geometric-mean speedup and a $3.88\times$ maximum on
depthwise Conv2D with asymmetric input and a square kernel. It improves 14 of 15 Level~2
fused-operator kernels, with a $1.17\times$ geometric mean and a
$2.55\times$ maximum on Matmul+Swish+Sum+GroupNorm. Across all 30 selected kernels,
AsmEvo improves 29, with a geometric-mean speedup of $1.35\times$. Despite the high improvement rate,
only 5.1\% of evaluated candidates become verified commits, indicating that
useful assembly rewrites are sparse.

The two splits exhibit different improvement profiles. Level~1 has a median speedup of
$1.31\times$; six kernels exceed $1.5\times$, spanning depthwise convolution,
normalization, activation, and matrix-multiplication operators. In addition to the
$3.88\times$ depthwise-convolution result, MinGPTNewGelu reaches $3.82\times$,
RMSNorm reaches $2.63\times$, and upper-triangular matrix multiplication reaches
$2.28\times$. The diversity of these operators indicates that the gains are not confined
to one instruction pattern or workload class.

Level~2 is more concentrated near parity: its median is $1.10\times$, and only
Matmul+Swish+Sum+GroupNorm exceeds $1.5\times$. The next-largest gains are
$1.32\times$ on ConvTranspose3d+AvgPool+Clamp+Softmax+Multiply and $1.30\times$
on ConvTranspose2d+Softmax+BiasAdd+Scaling+Sigmoid. One
GEMM+GroupNorm+Swish pipeline remains at $1.00\times$. This narrower distribution is
consistent with fused kernels exposing less local post-compilation headroom, while the
$2.55\times$ best case shows that substantial opportunities can still remain in individual
fused artifacts.

\begin{figure}[t]
  \centering
  \includegraphics[width=\columnwidth]{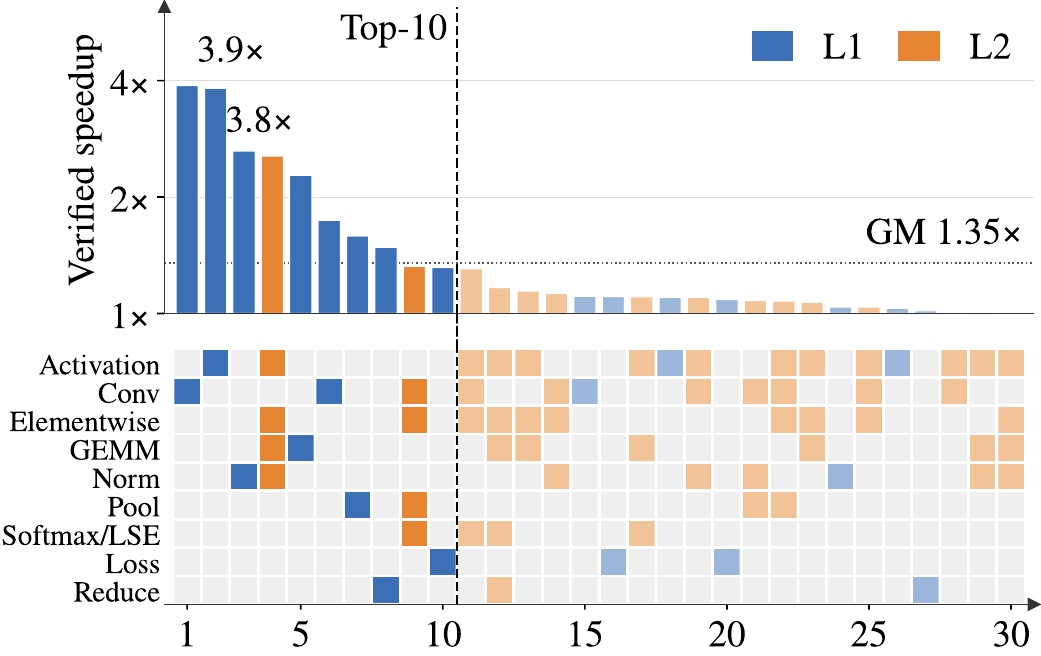}
  \caption{Verified AsmEvo speedups over the original code objects
  ($1.00\times$) for 30 selected KernelBench kernels, ranked by speedup. The lower panel
  shows each kernel's operator composition. All reported candidates pass functional-equivalence verification.}
  \label{fig:kernel-diversity}
\end{figure}

\subsubsection{Production Code Objects}
Table~\ref{tab:production-results} reports production kernels individually rather than as a
single aggregate because production binaries differ in launch semantics and oracle
requirements. For AITer code objects, AsmEvo improves all four retained kernels,
reaching a $1.09\times$ geometric mean speedup and a $1.31\times$ maximum speedup on the
FP8 blockscale MoE kernel. For Triton JIT compiled HSACOs from vLLM and SGLang, measured on MI300X, AsmEvo
improves all four assembly-level Triton kernels, reaching a $1.18\times$ geometric mean speedup and a
$1.34\times$ maximum speedup on the vLLM fused MoE kernel. These results show that the same
post-compilation harness applies beyond benchmark-generated kernels.

The production results also reveal that optimization headroom depends strongly on the
upstream artifact. The four AITer binaries contain one large gain and three comparatively
small gains, whereas three of the four Triton assembly results exceed $1.1\times$.
Because the kernels implement different operators and launch configurations, this comparison
does not isolate the upstream compiler as the cause; it instead demonstrates that AsmEvo can
recover useful opportunities from both vendor code objects and runtime-generated HSACOs.
We report only assembly-level optimization for Triton and binary-level optimization for
AITer; IR-level variants are excluded from all aggregates and claims.

\subsection{Case Study: Representative KernelBench Results}
The largest KernelBench gain occurs on
\path{84_conv_depthwise_2D_asymmetric_input_square_kernel}, which improves by
$3.88\times$. Two other Level~1 kernels show similarly substantial but structurally
different gains: \path{88_MinGPTNewGelu} reaches $3.82\times$, while
\path{36_RMSNorm_} reaches $2.63\times$. Upper-triangular matrix multiplication
improves by $2.28\times$. Together, these cases cover convolution, activation,
normalization, and structured matrix multiplication, illustrating why a single fixed
peephole rule would be insufficient for the evaluated set.

The strongest Level~2 case,
\path{37_Matmul_Swish_Sum_GroupNorm}, reaches $2.55\times$ despite combining
several operations in one compiled artifact. AsmEvo treats the complete kernel as the
acceptance unit: local assembly changes are rebuilt into the full code object and retained
only when the fused output passes the same differential oracle. This is important for fused
pipelines, where an apparently independent instruction sequence may affect a later reduction,
normalization, or activation stage.

\subsection{Case Study: Triton Production Kernels}
The strongest Triton assembly result is the vLLM
\path{fused_moe_kernel} at $1.343\times$. The vLLM
\path{gemm_a8w8_blockscale} kernel reaches $1.234\times$, showing that measurable
headroom remains even in a quantized block-scale GEMM emitted for a production serving
stack. On SGLang, \path{awq_gemm_kernel} reaches $1.133\times$, while
\path{sglang_fused_moe_lora_kernel} obtains a smaller but verified
$1.025\times$ gain. The range from $1.025\times$ to $1.343\times$ illustrates why
AsmEvo measures each rebuilt HSACO rather than assuming that a syntactically plausible
assembly transformation will improve every JIT-generated kernel.

For these kernels, validation uses captured production dispatches rather than reconstructed
synthetic arguments. The original kernarg layout, launch geometry, and referenced device
memory are replayed for both $K_0$ and the optimized object. This keeps the reported
assembly gains tied to the same pointer-rich MoE and quantized-GEMM states observed in the
serving frameworks.

\subsection{Case Study: AITer Code Objects}
AITer kernels provide a useful stress test because they are product-grade inference artifacts.
Although difficult to express at the source level after compilation, AsmEvo's gains arise
from narrow opportunities that remain visible in recovered AMDGCN assembly: scheduling around
long-latency memory operations, wait-counter placement, cache-hint and buffer-load variants,
and reduced conversion or address-generation work.

Across the AITer cases, the most promising edits are localized rather than structural. The
FP8 blockscale MoE kernel reaches \(1.31\times\), the FP8 causal FMHA kernel reaches
\(1.007\times\), and the two BF16 causal FMHA variants reach \(1.025\times\) and
\(1.049\times\). These attention and MoE kernels expose opportunities around routing cleanup,
pointer-rich access patterns, causal attention paths, and resource-pressure-sensitive
scheduling. Hoisting, interleaving, or tighter wait counters can reduce stalls without
changing the launcher ABI. Because synthetic launch reconstruction is often insufficient for
mixed-dtype, strided, or pointer-rich kernels, every accepted edit is replayed against a
captured production dispatch before its speedup is reported.

\subsection{Dispatch and End-to-End Validation}
AsmEvo rebuilds each accepted edit into a loadable AMDGPU code object that preserves the
original symbol, kernarg layout, descriptor-visible resources, and metadata contract. The
optimized artifact therefore follows the same host-side dispatch path as the original. For
KernelBench-derived HIP kernels, we replace the original MI308X object and run the same
benchmark harness.

For production libraries, we additionally validate the native runtime path. AITer
replacements pass its unit tests under the original launch configurations. For vLLM and
SGLang, we replace the cached Triton HSACO, disable recompilation, and run framework tests
against the replacement. These checks confirm that the measured speedups remain usable as
drop-in artifacts rather than only as replay-harness gains.

\section{Conclusion}

AsmEvo extends GPU kernel optimization to source-free AMDGPU code objects, using the original
binary as the behavioral oracle and admitting candidates only after ABI-preserving rebuild
and functional-equivalence verification. It improves 29 of 30 selected KernelBench kernels
($1.35\times$ geometric mean, $3.88\times$ maximum) and all eight production kernels from
AITer, vLLM, and SGLang. These results show that deployed binaries retain recoverable
low-level headroom and that agentic search can exploit it safely under deterministic gates.

\bibliography{aaai2027}

\end{document}